\documentclass{article}

\ifdefined\pdfminorversion
\fi

\usepackage{iclr2027_conference,times}
\usepackage{amsmath,amssymb}
\usepackage{algorithm}
\usepackage{algpseudocode}
\usepackage{booktabs}
\usepackage{graphicx}
\usepackage{microtype}
\usepackage{multirow}
\usepackage{placeins}
\usepackage{xcolor}
\usepackage{hyperref}
\usepackage{url}

\definecolor{SCULPTLink}{HTML}{1B4F72}
\definecolor{SCULPTCite}{HTML}{2E6B57}
\definecolor{SCULPTURL}{HTML}{1F6FB2}
\hypersetup{
  colorlinks=true,
  linkcolor=SCULPTLink,
  citecolor=SCULPTCite,
  urlcolor=SCULPTURL,
  breaklinks=true,
  bookmarksopen=true,
  bookmarksnumbered=true,
  bookmarksdepth=3,
  pdfdisplaydoctitle=true,
  pdfstartview=FitH,
  pdftitle={SCULPT: Subtractive Composition for 3D Part Generation},
  pdfauthor={Sikuang Li, Chen Yang, Jiemin Fang, Jiazhong Cen, Yuhe Wei, Wei Shen, Qi Tian},
  pdfsubject={Image-conditioned 3D object and part generation},
  pdfkeywords={3D generation, part generation, structured generation, subtractive composition}
}
\providecommand{\Description}[1]{}

\usepackage{hyperref}
\usepackage{url}
\usepackage{xspace}
\usepackage{graphicx}
\usepackage{wrapfig}
\usepackage{booktabs}
\usepackage{multirow}
\usepackage{caption}
\usepackage[dvipsnames]{xcolor}
\usepackage{hyperref}
\usepackage{marvosym} % provides \Letter

\makeatletter
\renewcommand*{\@fnsymbol}[1]{\ensuremath{\ifcase#1\or *\or \dagger\or
  \mbox{\Letter}\or \mathsection\or \mathparagraph\or \|\or **\or
  \dagger\dagger\or \ddagger\ddagger\else\@ctrerr\fi}}
\makeatother

\hypersetup{
  colorlinks=true,
  linkcolor=black,     %
  citecolor=black,     %
  urlcolor=RoyalBlue   %
}

\title{SCULPT: Subtractive Composition for\\ 3D Part Generation}

\author{
Sikuang Li$^1$\thanks{Equal Contribution.}\;\;\thanks{This work was done during an internship at Huawei.} \quad
Chen Yang$^2$\footnotemark[1] \quad
Jiemin Fang$^2$\thanks{Corresponding authors.}\quad\thanks{Project lead.}\quad\;
Jiazhong Cen$^1$ \\
\bf Yuhe Wei$^1$ \quad
Jichen Pang$^1$ \quad
Wei Shen$^1$\footnotemark[3] \quad\;
Qi Tian$^2$\footnotemark[3] \\
\\
$^1$Shanghai Jiao Tong University\quad
$^2$Huawei \\
\\
\texttt{\{uranusits, jiazhongcen, turtledoveden\}@sjtu.edu.cn}\\
\texttt{\{jiyuechenxing, wei.shen\}@sjtu.edu.cn}\\
\texttt{\{chenyang.res, jaminfong\}@gmail.com \quad tian.qi1@huawei.com} \\
\\
\url{https://sculpt-part.github.io/}
}

\iclrfinalcopy
\begin{document}

\maketitle
\lhead{}
\renewcommand{\headrulewidth}{0pt}

\begin{abstract}
Part-aware 3D generation aims to create digital assets that are coherent as complete objects while exposing structural parts for editing, material assignment, animation, and reuse. Existing methods impose this structure outside the native generation loop: segmentation-based methods partition an already generated shape, while additive methods synthesize parts from predefined layouts, boxes, or tokens and then reconcile them into a whole. The former preserves the generated geometry but fixes the object before part boundaries are determined; the latter exposes part cardinality but often leaves shared boundaries vulnerable to gaps, interpenetrations, and material discontinuities. In this paper, we propose \textbf{SCULPT}, a framework that addresses these challenges through \textit{subtractive composition}. Given a complete object represented in a structured 3D latent space, SCULPT iteratively applies a joint split predictor to generate one extracted part together with the remaining object. The predictor performs a coupled denoising process conditioned on both the image and the current 3D state, so the extracted part and updated remainder are generated together rather than reconciled after generation. The joint split predictor processes both outputs on the union of their native sparse 3D supports, allowing neighboring supports to overlap rather than imposing a disjoint voxel partition. The rollout ends when the remainder support becomes empty or reaches a fixed safety cap, allowing the number of generated parts to adapt to each object within that bound. Extensive experiments demonstrate state-of-the-art geometry on PartObjaverse while preserving strong complete-object reconstruction after part assembly. Results on four dataset images, one text-to-image-generated input, and one real-world photograph further show fine-grained textured part decomposition beyond the benchmark.
\end{abstract}

\begin{figure}[p]
  \centering
  \includegraphics[width=\linewidth,height=0.9\textheight,keepaspectratio]{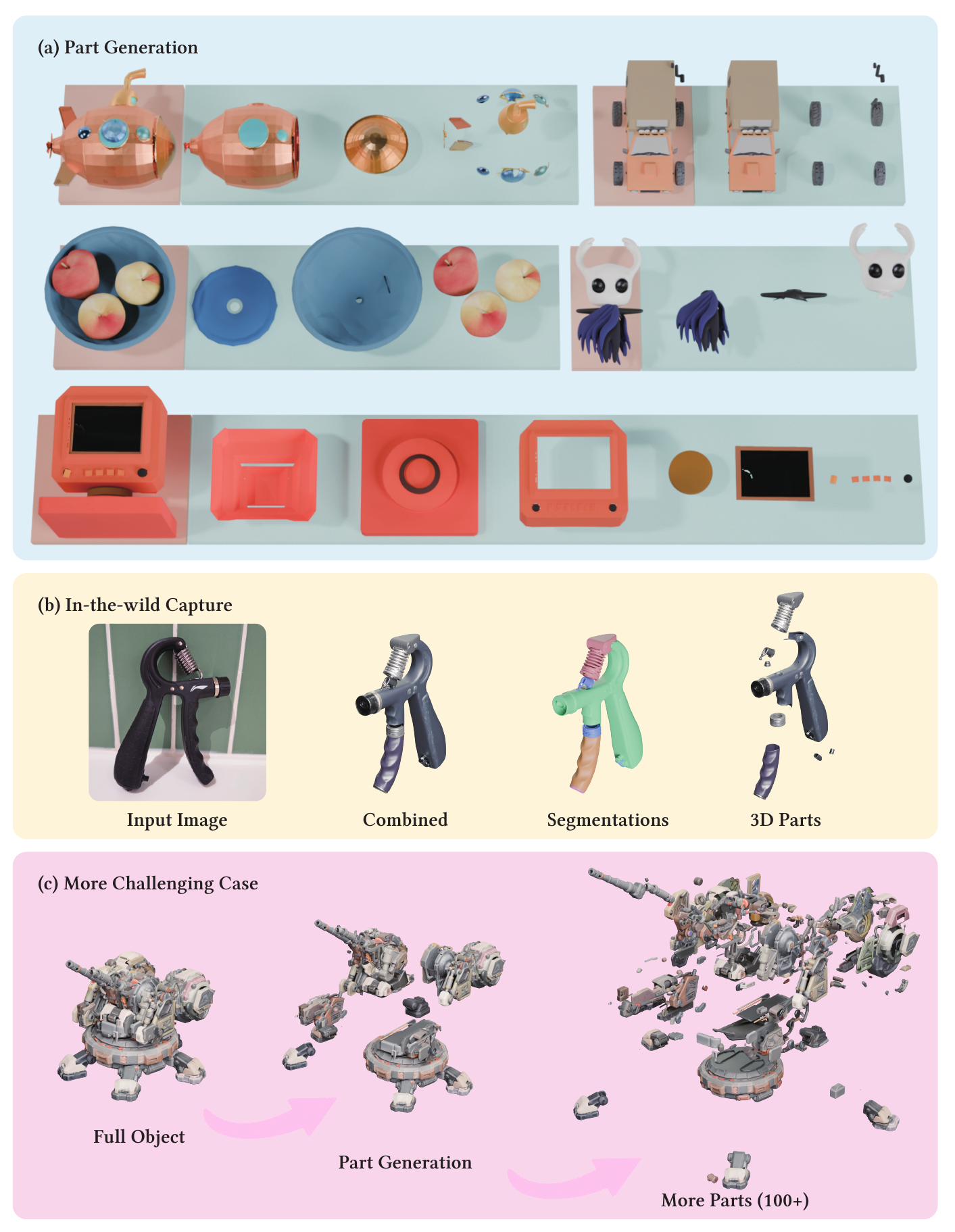}
  \caption{\textbf{SCULPT generates complete 3D objects together with coherent
  part structure.} Starting from an image-conditioned complete object, SCULPT
  subtractively extracts semantically meaningful components while retaining
  their alignment in the original object frame. (a)~Part generation: complete
  objects and their extracted parts across diverse object categories, with
  coherent geometry, boundaries, and appearance. (b)~In-the-wild capture: SCULPT generalizes well beyond its training
  distribution---given a single casually captured real photograph, it still
  produces a faithful combined object, semantic segmentations, and separated
  textured 3D parts.
  (c)~More challenging case: recursive subtractive decomposition---by applying
  the split operation to previously extracted parts, SCULPT decomposes a complex 3D
  asset into more than 100 fine-grained components.}
  \label{fig:teaser}
\end{figure}

\section{Introduction}
\label{sec:introduction}

In modern graphics pipelines, a 3D asset is rarely treated as a monolithic shape; rather, it is constructed as an assembly of meaningful parts. Whether for character animation, multi-material fabrication, or interactive content creation, artists routinely rig limbs, assign distinct materials to individual components, and edit sub-parts independently, yet they expect the assembled whole to remain visually coherent. While recent single-image 3D generative models such as TRELLIS~\citep{trellis}, TRELLIS.\,2~\citep{trellis2}, and Hunyuan3D 2.5~\citep{hunyuan2.5} excel at synthesizing detailed geometry and appearance from visual cues, they typically output a single, indivisible asset without considering the part structure. \emph{Part-aware 3D generation} asks these models to produce both: a coherent whole and an object-dependent number of semantic components that can be manipulated independently and reassembled. This variable output cardinality is central to the task.

Existing approaches handle part structure in two main ways, neither of which fully resolves this challenge. One line of work directly generates or reconstructs the complete part set. Part123~\citep{part123} and PartGen~\citep{partgen} derive components through image-space decomposition and per-part reconstruction, while OmniPart~\citep{omnipart} plans a variable-length box layout and jointly synthesizes its parts. More generally, direct part synthesis must represent the complete part set through image-space decisions, fixed slots, explicit layouts, or variable-length sequences. Fixed slots tie the output capacity to a chosen maximum, whereas variable-length representations must learn when and how many elements to emit. In either case, the object-dependent cardinality becomes part of a global output prediction. Segmentation-based methods take the complementary route: they defer the part decision until \emph{after} generation and label components on an already produced mesh or volume. Among these, 2D-guided methods such as SAMPart3D~\citep{sampart3d} transfer supervision from image foundation models, whereas native 3D methods such as PartField~\citep{partfield} and P3-SAM~\citep{p3sam} operate directly on meshes or 3D feature fields. Because these methods operate on a completed asset, they preserve its generated exterior geometry. A segmentation alone, however, partitions the existing surface rather than generatively completing the extracted components: it does not ask the generator to determine and complete both sides of a part boundary, including newly exposed contact surfaces and their geometry and materials. These complementary limitations motivate a formulation that retains the generator throughout decomposition without predicting the entire variable-cardinality part set at once.

Modern holistic 3D generators themselves provide the missing prior. Through whole-object generation, they learn recurring substructures---such as legs, wheels, handles, and panels---and how those substructures form coherent geometry and appearance. This suggests treating decomposition as a generation problem rather than only as recognition over a fixed asset. We therefore introduce \textbf{SCULPT}\footnote{The name \textbf{SCULPT} is inspired by sculpture as a subtractive process. Here, the analogy means that parts are derived from an initial whole: each step separates one meaningful component while keeping the remaining object.} and formulate part-aware generation as \emph{subtractive composition}. Instead of emitting the complete part set in one prediction, SCULPT repeatedly applies a fixed-signature operation: given the current object state, it jointly generates one extracted part and the remaining object. The remainder is then passed to the next split, and the same operation is reused until its predicted support becomes empty or the rollout reaches a fixed safety cap. By expressing cardinality through the rollout length rather than the dimensionality of a single prediction, subtractive composition alleviates the difficulty of modeling variable part cardinality. At the same time, generating each part together with its remainder allows their boundary to be determined during generation rather than imposed afterward.

To realize this generative split, we train a \emph{joint split predictor} that maps the current object state to the next extracted part and the updated remainder. The predictor accepts heterogeneous conditions: a 2D conditioning image and the current 3D state. Internally, it is implemented with decomposition flow transformer blocks, where a joint denoising branch initialized from the holistic generator produces the coupled part--remainder target and a remainder-control branch injects features from the current 3D state. In the initial step, the predictor operates on the complete structured latent of the object, yielding the first extracted part and the remaining object. Each subsequent iteration feeds the updated remaining latent back into the predictor to emit the next part. This native 3D coupling is particularly useful at part boundaries. Unlike 2D separation methods constrained by projected pixels, or voxelized partitions that snap boundaries to grid faces, SCULPT jointly processes the part and remainder on the union of their native sparse supports and allows them to share a non-empty interface shell. Boundary voxels can belong to both sides instead of being assigned by a hard voxel-face partition, while all outputs remain in the coordinate frame of the complete object. Fig.~\ref{fig:teaser} shows representative results.

In summary, our main contributions are:
\begin{itemize}
\item We formulate part-aware 3D generation as \emph{subtractive composition}, turning a variable-cardinality output into recurrent, fixed-signature part--remainder splits. The number of parts follows the rollout rather than a fixed set of output slots or a pre-specified layout.
\item We instantiate this formulation with a \emph{joint split predictor} built from decomposition flow transformer blocks. The predictor adapts the prior of a holistic image-to-3D generator through image-conditioned joint denoising and remainder-conditioned control, allowing each extracted part and its remainder to be generated in a shared trajectory rather than reconciled after independent synthesis.
\item We introduce native-support constraints for subtractive generation, including a sparse-support composition loss, inference-time support clipping, and empty-remainder termination. Together, these choices preserve coverage of the current object support and keep the recurrent state within the object being decomposed.
\item We evaluate SCULPT on a part-annotated Objaverse benchmark and image-driven generalization cases. SCULPT achieves state-of-the-art quantitative performance among part-structured generation and reconstruction baselines, including the best Chamfer distance at the part, semantic-group, and object levels, while producing coherent qualitative decompositions for four dataset images, one text-to-image-generated input, and one real-world photograph.
\end{itemize}

\section{Related Work}
\label{sec:related}
\subsection{Holistic 3D Object Generation}
\label{sec:related:object_generation}
3D object generation has seen rapid progress in recent years, with a variety of approaches for generating complete assets from text or images. Optimization-based methods~\citep{dreamfusion,gaussiandreamer,magic3d,mvdream} use 2D or multi-view diffusion priors~\citep{ddpm,ddim} to optimize 3D representations. Direct 3D diffusion models generate point clouds or implicit functions~\citep{pointe,shape}, while feed-forward and sparse-view reconstruction pipelines improve efficiency and view consistency~\citep{wonder3d,gaussianobject,lrm,lgm,instantmesh}. More recent systems, including TRELLIS~\citep{trellis}, TRELLIS.\,2~\citep{trellis2}, Hunyuan3D 2.5~\citep{hunyuan2.5}, UniLat3D~\citep{unilat3d}, and TIGON~\citep{tigon}, build on native 3D latent representations to support high-quality asset generation. Beyond single objects, generative pipelines have also been scaled to full scenes, where layouts or language arrange assets into coherent environments~\citep{scenecraft,holodeck}.

\subsection{3D Part Segmentation}
\label{sec:related:part_segmentation}
3D part segmentation methods decompose an object into semantically meaningful components, providing a basis for part-level understanding and manipulation. Early approaches segment shapes into geometric primitives with learned point-wise features~\citep{hpnet}, often trained on part-annotated datasets such as PartNet~\citep{partnet}. Recent methods obtain part cues from 2D foundation models~\citep{dinov2,dinov3,sam} or operate with native 3D backbones~\citep{pointnetpp,ptv3}, including PartSLIP, Point-SAM, PartSAM, SAMPart3D, and P3-SAM~\citep{partslip,pointsam,partsam,sampart3d,p3sam}. Their features are clustered, classified, or decoded to produce part labels. Segmentation has also been lifted to reconstructed scene representations: Gaussian Grouping~\citep{gaussiangrouping} attaches identity encodings to 3D Gaussians supervised by 2D masks, enabling open-world scene segmentation and editing. Because these methods operate on a fixed asset, they preserve the generated geometry and provide a strong reference for part-level decomposition. Their output, however, is a partition of an existing shape rather than a generative process, so the decomposition does not directly update the object state as parts are extracted.

\subsection{Part-Structured 3D Generation and Reconstruction}
\label{sec:related:part_generation}
Part-structured 3D generation and reconstruction make part structure native to the synthesis or recovery process. Early structure-aware methods represent shapes as part hierarchies and learn structural variations for generation and editing~\citep{structedit}. Building on modern generative backbones, some methods use multi-view generation or image-space part cues to recover complete semantic assemblies, including PartGen~\citep{partgen}, HoloPart~\citep{holopart}, and Part123~\citep{part123}. Other systems model part-structured objects through dual volume packing~\citep{partpacker}, compositional latent diffusion in PartCrafter~\citep{partcrafter}, semantic decoupling and structural cohesion in OmniPart~\citep{omnipart}, contextual part latents in CoPart~\citep{copart}, bounding-box-prompted shape decomposition in X-Part~\citep{xpart}, or full-resolution per-part synthesis in FullPart~\citep{fullpart}. Part structure has further been exposed through alternative interfaces: BANG~\citep{bang} separates an asset by generating exploded-view dynamics with spatial prompts, and Part-X-MLLM~\citep{partxmllm} plans part-level boxes and edit commands with a language frontend that drives downstream geometry modules. These methods expose part cardinality and semantics during synthesis or reconstruction, giving them direct control over the structure being produced. Maintaining coherence among generated parts remains an explicit design problem, addressed through packing, joint modeling, structural conditioning, or completion. Closest to our formulation, UniPart~\citep{unipart} likewise taps the part-aware priors that emerge in whole-object generative learning, yet it realizes decomposition as a latent segmentation stage followed by per-part diffusion. AutoPartGen~\citep{autopartgen} generates parts autoregressively with automatic termination, but each part is added by conditioning on the whole object and previously generated parts, so no explicit remainder state constrains what remains to be generated, and its geometry-only latents do not model appearance. In contrast, SCULPT carries an explicit remaining-object state through every split and denoises each part jointly with that remainder on the union of their native sparse supports, keeping boundaries, geometry, and materials consistent as the decomposition proceeds.

\subsection{Conditional Control and Joint Generation}
\label{sec:related:conditional_control}
Conditional control studies how to steer pretrained generative models with additional signals while retaining their learned priors. ControlNet~\citep{controlnet} adds trainable branches to text-to-image diffusion models for spatial conditions such as edges, depth, or pose, and adapter-based methods such as T2I-Adapter~\citep{t2iadapter} and IP-Adapter~\citep{ipadapter} inject spatial controls or image prompts with lightweight modules. JointNet~\citep{jointnet} and JointDiT~\citep{jointdit} further show that diffusion backbones can be extended to coupled outputs such as RGB-depth generation, while text-image conditioned 3D generation~\citep{tigon} explores multimodal control by combining visual exemplars with textual specifications. These works demonstrate the potential of conditioning and joint generation to steer pretrained models toward more complex outputs, but they do not directly address recurrent 3D part decomposition conditioned on an evolving remaining-object state.

\section{Method}
\label{sec:method}

\begin{figure}[t]
  \centering
  \includegraphics[width=\linewidth]{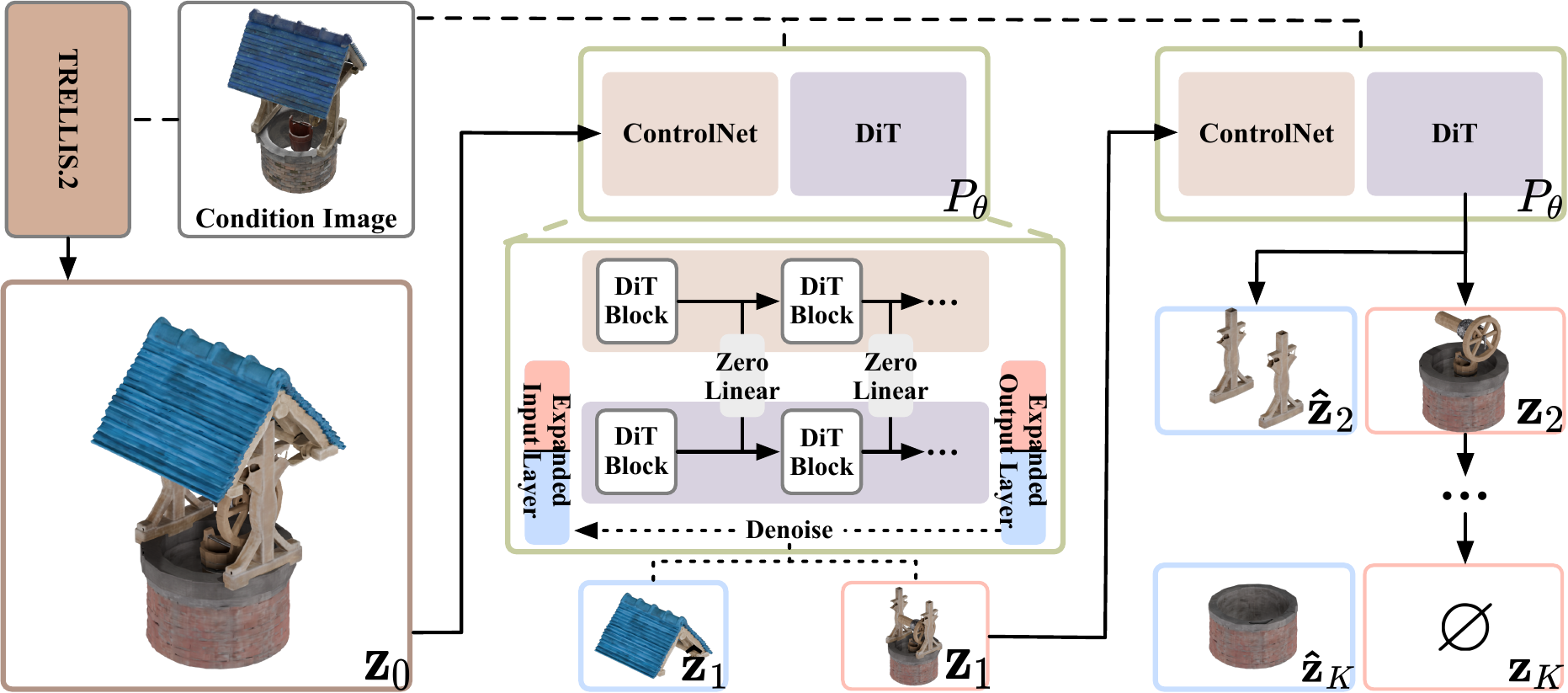}
  \caption{\textbf{SCULPT turns whole-object generation into recurrent
  part--remainder prediction.} We first use TRELLIS.\,2 mapping the conditioning image
  $\mathcal{I}$ to a complete structured latent $\boldsymbol{z}_0$. At split
  $i$, the image- and remainder-conditioned predictor $P_\theta$ jointly
  denoises an extracted part $\hat{\boldsymbol{z}}_i$ (blue) and the updated
  remainder $\boldsymbol{z}_i$ (red); only the remainder state is carried into
  the next split, until its sparse support becomes empty or the rollout reaches
  its safety cap. The inset expands $P_\theta$: a ControlNet branch encodes the
  current remainder and injects zero-initialized residuals into the DiT blocks.}
  \Description{SCULPT generates a complete structured latent and recurrently
  predicts one part together with the remaining object until the remainder is
  empty.}
  \label{fig:method_overview}
\end{figure}

\subsection{Overview}
\label{sec:method:overview}
\label{sec:method:formulation}

Given a conditioning image $\mathcal{I}$, SCULPT generates an ordered,
variable-length collection of 3D parts. Each part latent decodes to a textured
mesh with geometry and material in the coordinate frame of the complete object,
so the decoded parts can be combined directly into one asset. We call
the process \emph{subtractive composition}: starting from a complete object,
the model repeatedly separates one part and carries the remaining object to the
next step.

SCULPT uses the pretrained image-conditioned TRELLIS.\,2 generator
$G_\phi$~\citep{trellis2} to obtain the complete-object latent
$\boldsymbol{z}_0$. Its core module is a \emph{joint split predictor}
$P_\theta$. Let $\boldsymbol{z}_{i-1}$ be the current remaining-object latent
before split $i$. The predictor takes this latent together with the image and
returns the next part $\hat{\boldsymbol{z}}_i$ and the updated remainder
$\boldsymbol{z}_i$:
\begin{align}
  \boldsymbol{z}_0 &= G_\phi(\mathcal{I}),
  \label{eq:initial_whole}\\
  (\hat{\boldsymbol{z}}_i,\boldsymbol{z}_i)
  &=P_\theta(\boldsymbol{z}_{i-1},\mathcal{I}).
  \label{eq:joint_split}
\end{align}
Here and below, a hatted quantity denotes the extracted part and the
corresponding unhatted quantity denotes the remainder. The extracted latent is
stored, while the remainder is used as the input to the next split. Repeating
the same two-output prediction allows different objects to produce different
numbers of parts, up to the fixed rollout cap. Fig.~\ref{fig:method_overview}
summarizes this process.

The remainder of this section follows the pipeline. We first review the
structured 3D latent used for complete objects, parts, and remainders. We then
describe the subtractive training sequences, the joint predictor and its
objective, and the recurrent inference procedure.

\subsection{Preliminaries: Structured Latent Backbone}
\label{sec:method:preliminaries}
\label{sec:method:structured_latent}

Jointly predicting a part and a remainder requires both outputs to encode
geometry and material in the coordinate system of the complete object. We therefore
use the structured latent representation of TRELLIS.\,2~\citep{trellis2} for
every complete object, part, and remainder:
\begin{equation}
  \boldsymbol{z}
  =(\boldsymbol{z}^{v},\boldsymbol{z}^{g},\boldsymbol{z}^{m}),
  \qquad s\in\{v,g,m\},
  \label{eq:structured_latent}
\end{equation}
where $v$, $g$, and $m$ index the sparse-structure, geometry, and material
stages. The sparse-structure latent determines an active O-Voxel support
$\mathcal{O}\in\{0,1\}^{N^3}$ on an O-Voxel grid, the backbone's sparse voxel
representation, at resolution $N$. The geometry stage generates features on
this support, and the material stage generates features aligned with the
geometry. We use $\cup$, $\cap$, and $\varnothing$ for elementwise operations
on these supports. The decoder $\mathcal{D}$ converts the three latent
components into a textured mesh in a shared $[-1,1]^3$ object frame. Because
parts and remainders use this same frame, they do not require independent
normalization or registration before composition.

\begin{figure}[t]
  \centering
  \includegraphics[width=\linewidth]{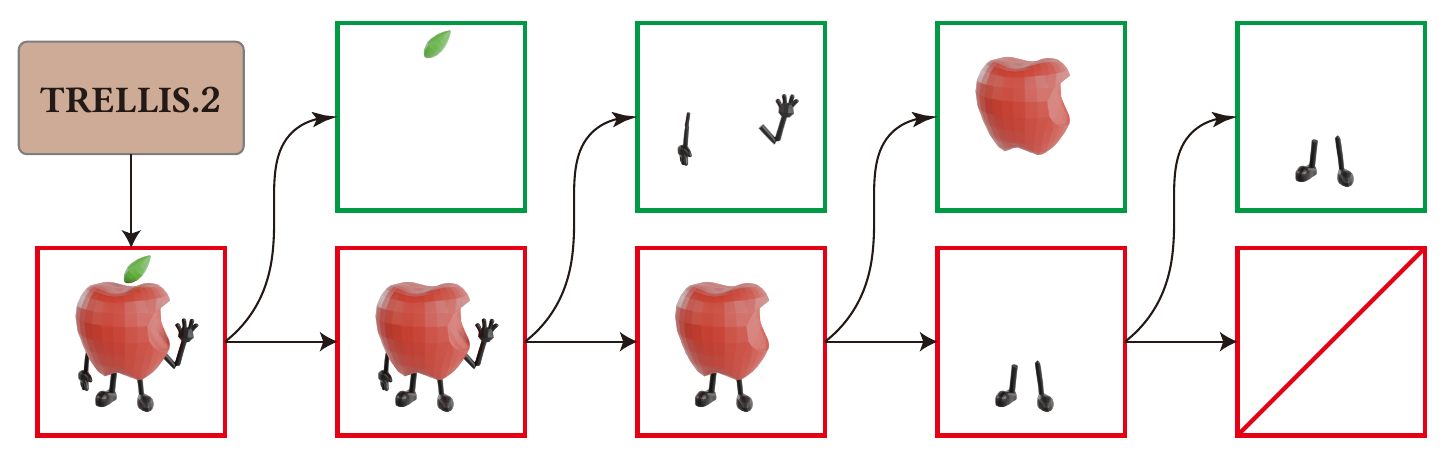}
  \caption{\textbf{A decomposition unfolds as a sequence of part--remainder
  transitions.} Starting from the TRELLIS.\,2 whole-object state (red), each
  split emits one stored part (green) and carries only the updated remainder
  forward. In this example, the leaf, arms, body, and legs are removed
  successively; the rollout terminates when the remainder becomes empty.}
  \Description{A complete object is recurrently split into an extracted part
  and an updated remainder until no remainder support remains.}
  \label{fig:subtractive_sequence}
\end{figure}

Each of the three TRELLIS.\,2 stages is a rectified-flow generator. For a clean
target $\boldsymbol{x}_0$, noise $\boldsymbol{\epsilon}$, and time
$t\in[0,1]$, the forward path is
\begin{equation}
  \boldsymbol{x}(t)
  =(1-t)\boldsymbol{x}_0+t\boldsymbol{\epsilon},
  \qquad
  \dot{\boldsymbol{x}}(t)=\boldsymbol{\epsilon}-\boldsymbol{x}_0.
  \label{eq:flow_path}
\end{equation}
A velocity network $\boldsymbol{v}_\theta$ is trained by conditional flow
matching to predict this vector field,
\begin{equation}
  \mathcal{L}_{\mathrm{CFM}}(\theta)
  =\mathbb{E}_{t,\boldsymbol{x}_0,\boldsymbol{\epsilon}}
  \left\|
  \boldsymbol{v}_\theta(\boldsymbol{x}(t),t)
  -(\boldsymbol{\epsilon}-\boldsymbol{x}_0)
  \right\|_2^2.
  \label{eq:cfm}
\end{equation}
SCULPT retains this stage order and flow parameterization, but changes the clean
target from one latent to a coupled part--remainder pair and conditions the
velocity network on the current remainder. Fig.~\ref{fig:subtractive_sequence}
gives a concrete view of the resulting recurrence: each split stores one part
and carries the updated remainder into the next step.

\subsection{Subtractive Training Sequences}
\label{sec:method:data_curation}

Training the split predictor requires supervision for both outputs of every
step. We construct this supervision from part-structured 3D assets that provide
a complete mesh and its part meshes. The complete mesh and all of its parts are
jointly normalized to the shared object frame; individual parts are never
rescaled. We then sort the parts lexicographically by their centroids along the
$z$--$x$--$y$ axes, giving one deterministic extraction order
per asset.

Let $\mathcal{M}_0$ be a complete mesh with $K$ ordered parts, let
$\hat{\mathcal{M}}_i$ be the part extracted at step $i$, and let
$\mathcal{M}_i$ be the mesh remaining after the first $i$ parts have been
extracted. These meshes satisfy
\begin{align}
  \mathcal{M}_{i-1}
  &=\hat{\mathcal{M}}_i\oplus\mathcal{M}_i,
  &&i=1,\ldots,K,
  \label{eq:mesh_recurrence}\\
  \mathcal{T}^{\mathcal{M}}_i
  &=(\mathcal{M}_{i-1};\hat{\mathcal{M}}_i,\mathcal{M}_i),
  &&i=1,\ldots,K,
  \label{eq:data_mesh_tuple}\\
  \mathcal{M}_K&=\varnothing,
  &\mathcal{S}(\mathcal{M}_0)
  &=\{\mathcal{T}^{\mathcal{M}}_i\}_{i=1}^{K},
  \label{eq:mesh_sequence}
\end{align}
where $\oplus$ denotes mesh union in the shared frame. Thus, one annotated asset
provides $K$ examples of splitting a current object into one part and a new
remainder. The final tuple has an empty target remainder and provides the
supervision used to learn termination.

We render the conditioning image $\mathcal{I}$ from the complete mesh. Each
extracted part $\hat{\mathcal{M}}_i$ and remainder $\mathcal{M}_i$ is then
converted independently to an O-Voxel in the common frame and encoded as
$\hat{\boldsymbol{z}}_i$ and $\boldsymbol{z}_i$. Independent conversion allows
the part support $\hat{\mathcal{O}}_i$ and remainder support $\mathcal{O}_i$ to
overlap at their contact region instead of forcing a disjoint voxel partition.
For the terminal state, the sparse-structure target is an all-empty occupancy
grid, while the geometry and material targets are zero-length sparse tensors.
The resulting latent training tuple is
\begin{equation}
  \mathcal{T}_i
  =(\mathcal{I},\boldsymbol{z}_{i-1};
  \hat{\boldsymbol{z}}_i,\boldsymbol{z}_i),
  \qquad i=1,\ldots,K.
  \label{eq:data_latent_tuple}
\end{equation}
This construction matches the input and outputs of
Eq.~\ref{eq:joint_split}, keeps all targets geometrically aligned, and uses the
same predictor interface for nonempty and terminal remainders. During training,
$\boldsymbol{z}_{i-1}$ is encoded from the annotated remainder mesh; during
inference, it is the remainder predicted by the preceding split.

\subsection{Decomposition Flow Transformer}
\label{sec:method:decomposition_flow_transformer}
\label{sec:method:branch}

The split predictor must use two complementary conditions: the image describes
the complete asset, while the current remainder specifies which 3D content is
still present at the current step. We implement $P_\theta$ as three stage-wise
\emph{decomposition flow transformers},
$P_\theta=\{P_\theta^v,P_\theta^g,P_\theta^m\}$, following the sparse-structure,
geometry, and material order of the backbone.

\paragraph{Stage-wise joint prediction.}
For stage $s$, we pack the extracted-part and remainder targets into
$\boldsymbol{y}_i^s=(\hat{\boldsymbol{z}}_i^s,\boldsymbol{z}_i^s)$. One split
runs the three predictors in order:
\begin{align}
  (\hat{\boldsymbol{z}}_i^v,\boldsymbol{z}_i^v)
  &=P_\theta^v(\boldsymbol{z}_{i-1}^v,\mathcal{I}),
  \label{eq:stage_sparse}\\
  (\hat{\boldsymbol{z}}_i^g,\boldsymbol{z}_i^g)
  &=P_\theta^g(\boldsymbol{z}_{i-1}^g,\mathcal{I};
  \hat{\mathcal{O}}_i,\mathcal{O}_i),
  \label{eq:stage_geometry}\\
  (\hat{\boldsymbol{z}}_i^m,\boldsymbol{z}_i^m)
  &=P_\theta^m(\boldsymbol{z}_{i-1}^m,\mathcal{I};
  \hat{\boldsymbol{z}}_i^g,\boldsymbol{z}_i^g).
  \label{eq:stage_material}
\end{align}
At the sparse-structure stage, $\boldsymbol{y}_i^v$ is represented as a
two-channel dense tensor on the $N^3$ grid. At the geometry and material
stages, the part and remainder can occupy different sparse coordinates. We
place their features on the union support
$\hat{\mathcal{O}}_i\cup\mathcal{O}_i$, retain masks for the two destinations,
jointly denoise the packed features, and route the two outputs back to their
respective supports. The two outputs therefore share the same denoising
computation and image context at every stage.

\paragraph{Remainder-conditioned blocks.}
Each stage predictor contains a joint denoising branch and a parallel
remainder-control branch. Let $\boldsymbol{h}_{\ell,i}^s$ be the packed feature
in the joint branch at block $\ell$, let $\boldsymbol{c}_{\ell,i}^s$ be the
control feature, and let $\boldsymbol{e}_{\mathcal I}$ and
$\boldsymbol{e}_t$ be the image and flow-time embeddings. We denote the
control and joint transformer blocks by $C_\ell^s$ and $B_\ell^s$, and their
residual projection by $Z_\ell^s$. A block updates the two branches as
\begin{align}
  \boldsymbol{c}_{\ell+1,i}^s
  &=C_\ell^s\!\left(\boldsymbol{c}_{\ell,i}^s,
  \boldsymbol{z}_{i-1}^s;
  \boldsymbol{e}_{\mathcal I},\boldsymbol{e}_t\right),
  \label{eq:control_block}\\
  \boldsymbol{h}_{\ell+1,i}^s
  &=B_\ell^s\!\left(\boldsymbol{h}_{\ell,i}^s;
  \boldsymbol{e}_{\mathcal I},\boldsymbol{e}_t\right)
  +Z_\ell^s\!\left(\boldsymbol{c}_{\ell+1,i}^s\right).
  \label{eq:joint_block}
\end{align}
The joint blocks $B_\ell^s$ are initialized from the corresponding
TRELLIS.\,2 checkpoint. The residual projections $Z_\ell^s$ are initialized to
zero, and both branches are then trained for the split task. This design starts
from the pretrained image-to-3D generator while adding the current 3D remainder
as a block-wise condition.

\paragraph{Training objectives.}
\label{sec:method:objectives}
For each stage $s$, the packed target follows the rectified-flow path
\begin{equation}
  \boldsymbol{y}_i^s(t)
  =(1-t)\boldsymbol{y}_i^s+t\boldsymbol{\epsilon}.
  \label{eq:joint_flow_path}
\end{equation}
We train the stage-wise velocity predictor with
\begin{equation}
  \mathcal{L}_{\mathrm{flow}}^s
  =\mathbb{E}
  \left[
  \left\|
  \boldsymbol{v}_\theta^s(\boldsymbol{y}_i^s(t),t,
  \mathcal{I},\boldsymbol{z}_{i-1}^s)
  -(\boldsymbol{\epsilon}-\boldsymbol{y}_i^s)
  \right\|_2^2
  \right],
  \label{eq:joint_flow_loss}
\end{equation}
where the expectation is over latent training tuples, flow times, and noise.
The packed target fixes the roles of the two outputs, while joint denoising lets
their predictions interact throughout the flow trajectory.

The flow loss supervises the two target latents but does not directly compare
their combined sparse support with the input remainder. We therefore add a
composition loss at the sparse-structure stage. From the predicted velocity, we
first estimate the clean packed state,
\begin{equation}
  \tilde{\boldsymbol{y}}_{i,0}^v(t)
  =\boldsymbol{y}_i^v(t)-t\,
  \boldsymbol{v}_\theta^v(\boldsymbol{y}_i^v(t),t,
  \mathcal{I},\boldsymbol{z}_{i-1}^v).
  \label{eq:clean_support_estimate}
\end{equation}
After unpacking its two occupancy-logit channels and applying a sigmoid, we
obtain part and remainder occupancy probabilities
$(\hat{\boldsymbol{p}}_i,\boldsymbol{p}_i)$. Their differentiable union is
\begin{equation}
  \boldsymbol{u}_i
  =\mathbf{1}-(\mathbf{1}-\hat{\boldsymbol{p}}_i)
  \odot(\mathbf{1}-\boldsymbol{p}_i).
  \label{eq:soft_union}
\end{equation}
where $\odot$ denotes elementwise multiplication. We compare this union with
the support $\mathcal{O}_{i-1}$ of the input remainder using voxel-wise binary
cross-entropy, where $q$ indexes grid voxels. The sparse-stage objective adds
this term with weight $\lambda_{\mathrm{comp}}$:
\begin{align}
  \mathcal{L}_{\mathrm{comp}}
  &=-\frac{1}{N^3}\sum_q\left[
  \mathcal{O}_{i-1,q}\log u_{i,q}
  +(1-\mathcal{O}_{i-1,q})\log(1-u_{i,q})
  \right],
  \label{eq:composition_loss}\\
  \mathcal{L}^{v}
  &=\mathcal{L}_{\mathrm{flow}}^{v}
  +\lambda_{\mathrm{comp}}\mathcal{L}_{\mathrm{comp}},
  \qquad
  \mathcal{L}^{g}=\mathcal{L}_{\mathrm{flow}}^{g},
  \qquad
  \mathcal{L}^{m}=\mathcal{L}_{\mathrm{flow}}^{m}.
  \label{eq:dft_objective}
\end{align}
The three stages are trained separately with their corresponding objectives.
The composition term encourages the predicted part and remainder together to
cover the current input support. It constrains their union, not their
intersection, so voxels may be active in both outputs. We call such overlapping
boundary voxels an \emph{interface shell}; they avoid imposing a hard voxel-face
partition between two parts that meet in 3D.

\subsection{Subtractive Composition}
\label{sec:method:subtractive_composition}
\label{sec:method:decomposition}
\label{sec:method:stopping}

At inference time, predictions from one split become the input to the next.
To keep this recurrent state within the current object, we convert the two
sparse-stage probability maps into hard supports with a fixed occupancy
threshold $\tau_{\mathrm{occ}}$ and clip both supports to
$\mathcal{O}_{i-1}$:
\begin{equation}
  \hat{\mathcal{O}}_i
  =\mathbf{1}[\hat{\boldsymbol{p}}_i>\tau_{\mathrm{occ}}]
  \cap\mathcal{O}_{i-1},
  \qquad
  \mathcal{O}_i
  =\mathbf{1}[\boldsymbol{p}_i>\tau_{\mathrm{occ}}]
  \cap\mathcal{O}_{i-1}.
  \label{eq:inference_clip}
\end{equation}
Here $\mathbf{1}[\cdot]$ is the indicator function. We mask the predicted sparse latents to these clipped supports before running
the geometry and material stages. The updated remainder latent
$\boldsymbol{z}_i$ is then carried to the next split, while
$\hat{\boldsymbol{z}}_i$ is stored as an output part. Algorithm~\ref{alg:subtractive_composition}
gives the complete procedure.

\begin{algorithm}[t]
\caption{\textbf{SCULPT recurrent inference.} Each split predicts and clips an
extracted part and updated remainder. The rollout stops at an empty remainder
or after $K_{\max}=24$ splits, retaining a nonempty capped remainder before
decoding and assembly.}
\label{alg:subtractive_composition}
\begin{algorithmic}[1]
\Require image $\mathcal{I}$; whole-object generator $G_\phi$; trained split
predictors $P_\theta^v,P_\theta^g,P_\theta^m$; decoder $\mathcal{D}$; threshold
$\tau_{\mathrm{occ}}$; cap $K_{\max}=24$
\Ensure ordered output sequence $\mathcal{Z}_{\mathrm{out}}$ and assembled asset
\State $\boldsymbol{z}_0\gets G_\phi(\mathcal{I})$ and extract
$\mathcal{O}_0$ from $\boldsymbol{z}_0^v$
\State $\mathcal{Z}_{\mathrm{out}}\gets[\,]$
\For{$i=1,\ldots,K_{\max}$}
  \State $(\hat{\boldsymbol{z}}_i^v,\boldsymbol{z}_i^v)
  \gets P_\theta^v(\boldsymbol{z}_{i-1}^v,\mathcal{I})$
  \State Threshold and mask both sparse latents, yielding the clipped supports
  in Eq.~\ref{eq:inference_clip}
  \State $(\hat{\boldsymbol{z}}_i^g,\boldsymbol{z}_i^g)
  \gets P_\theta^g(\boldsymbol{z}_{i-1}^g,\mathcal{I};
  \hat{\mathcal{O}}_i,\mathcal{O}_i)$
  \State $(\hat{\boldsymbol{z}}_i^m,\boldsymbol{z}_i^m)
  \gets P_\theta^m(\boldsymbol{z}_{i-1}^m,\mathcal{I};
  \hat{\boldsymbol{z}}_i^g,\boldsymbol{z}_i^g)$
  \State Assemble $\hat{\boldsymbol{z}}_i$ and $\boldsymbol{z}_i$ from their
  $v,g,m$ components
  \State $\operatorname{append}(\mathcal{Z}_{\mathrm{out}},
  \hat{\boldsymbol{z}}_i)$; $K\gets i$
  \If{$\mathcal{O}_i=\varnothing$}
    \State \textbf{break}
  \EndIf
\EndFor
\If{$\mathcal{O}_K\neq\varnothing$}
  \State $\operatorname{append}(\mathcal{Z}_{\mathrm{out}},
  \boldsymbol{z}_K)$
  \Comment{retain capped remainder}
\EndIf
\State \Return $\mathcal{Z}_{\mathrm{out}}$ and
$\displaystyle\bigcup_{\boldsymbol{z}\in\mathcal{Z}_{\mathrm{out}}}
\mathcal{D}(\boldsymbol{z})$
\end{algorithmic}
\end{algorithm}

\label{sec:method:properties}
Clipping gives a non-expanding sequence of supports,
\begin{equation}
  \hat{\mathcal{O}}_i\subseteq\mathcal{O}_{i-1},
  \qquad
  \mathcal{O}_i\subseteq\mathcal{O}_{i-1}
  \subseteq\cdots\subseteq\mathcal{O}_0.
  \label{eq:support_invariant}
\end{equation}
The rollout stops when the remainder support is empty or when it reaches the
fixed cap $K_{\max}=24$. Equivalently, its final split index is
\begin{equation}
  K=\min\!\left(
  \{i\leq K_{\max}:|\mathcal{O}_i|=0\}
  \cup\{K_{\max}\}
  \right).
  \label{eq:stop_index}
\end{equation}
If an empty support is predicted first, the output contains the $K$ extracted
part latents. If the cap is reached while the remainder is nonempty, that
remainder is retained as one additional output instead of being discarded.
Finally, each output latent is decoded independently by $\mathcal{D}$, and the
decoded meshes are united in the shared object frame without per-part
rescaling, snapping, or registration. The recurrent formulation therefore
keeps a fixed two-output prediction interface while allowing the number of
generated parts to follow the rollout length. At the same time, clipping keeps
every predicted state inside the initial object support, and retaining a
nonempty capped remainder prevents unresolved geometry from being silently
discarded.

\section{Experiments}
\label{sec:experiments}

\begin{table}[t]
  \centering
  \normalsize
  \caption{\textbf{Geometry comparison on PartObjaverse under matched
  single-image conditioning.} Part-level scores use Hungarian-matched
  instances; semantic-group scores union parts sharing a semantic label;
  object-level scores use the assembled union. Lower CD and higher F1 are
  better. Bold and underlined entries mark the best and second-best available
  values within each level, respectively; ``--'' denotes an unavailable
  measurement.}
  \label{tab:eval_generation}
  \begin{tabular*}{\linewidth}{@{\extracolsep{\fill}}lccc@{}}
    \toprule
    Method & CD$\downarrow$ & F1@.1$\uparrow$ & F1@.05$\uparrow$ \\
    \midrule
    \multicolumn{4}{@{}l}{\textit{Part level}} \\
    Part123 & 0.0188 & 0.8133 & 0.6279 \\
    OmniPart & 0.0136 & 0.8612 & 0.7025 \\
    TRELLIS+PartField & 0.0127 & 0.8750 & 0.7169 \\
    TRELLIS+SAM3D & 0.0278 & 0.7410 & 0.5520 \\
    TRELLIS+PartField+HoloPart & 0.0137 & 0.8521 & 0.6862 \\
    TRELLIS.2+PartField & \underline{0.0115} & \textbf{0.8897} & \underline{0.7554} \\
    TRELLIS.2+SAM3D & 0.0531 & 0.5323 & 0.3283 \\
    TRELLIS.2+PartField+HoloPart & 0.0453 & 0.6016 & 0.3966 \\
    \textbf{SCULPT (ours)} & \textbf{0.0107} & \underline{0.8858} & \textbf{0.7599} \\
    \midrule
    \multicolumn{4}{@{}l}{\textit{Semantic-group level}} \\
    Part123 & 0.0194 & 0.8105 & 0.6299 \\
    OmniPart & 0.0135 & 0.8623 & 0.7060 \\
    TRELLIS+PartField & 0.0130 & 0.8734 & 0.7197 \\
    TRELLIS+SAM3D & 0.0261 & 0.7483 & 0.5583 \\
    TRELLIS+PartField+HoloPart & 0.0139 & 0.8510 & 0.6879 \\
    TRELLIS.2+PartField & \underline{0.0117} & \textbf{0.8903} & \underline{0.7595} \\
    TRELLIS.2+SAM3D & 0.0505 & 0.5528 & 0.3486 \\
    TRELLIS.2+PartField+HoloPart & 0.0441 & 0.6156 & 0.4106 \\
    \textbf{SCULPT (ours)} & \textbf{0.0107} & \underline{0.8851} & \textbf{0.7614} \\
    \midrule
    \multicolumn{4}{@{}l}{\textit{Object level}} \\
    Part123 & 0.0126 & 0.8229 & 0.5739 \\
    OmniPart & 0.0032 & 0.9690 & 0.8732 \\
    TRELLIS+PartField & 0.0037 & 0.9634 & 0.8563 \\
    TRELLIS+SAM3D & -- & -- & -- \\
    TRELLIS+PartField+HoloPart & 0.0034 & 0.9654 & 0.8639 \\
    TRELLIS.2+PartField & \underline{0.0021} & \underline{0.9793} & \underline{0.9119} \\
    TRELLIS.2+SAM3D & -- & -- & -- \\
    TRELLIS.2+PartField+HoloPart & 0.0332 & 0.6474 & 0.4345 \\
    \textbf{SCULPT (ours)} & \textbf{0.0020} & \textbf{0.9839} & \textbf{0.9212} \\
    \bottomrule
  \end{tabular*}
\end{table}

\subsection{Experimental Setup}
\label{sec:exp:settings}

\paragraph{Training data.}
PartVerse-XL~\citep{fullpart} provides the supervision required by our
subtractive formulation: each asset contains human-refined part meshes aligned
with its complete mesh. The dataset is curated from
Objaverse-XL~\citep{objaverseXL}. We remove assets with excessive component
counts, extremely small components, missing PBR materials, or low-quality
geometry and texture. After filtering, the training set contains 37,425 objects
and 330,455 supervised part--remainder splits. To prevent evaluation leakage,
we also remove every training asset whose SHA-256 identifier matches an
evaluation mesh.

\begin{figure}[thbp]
  \centering
  \includegraphics[width=\linewidth]{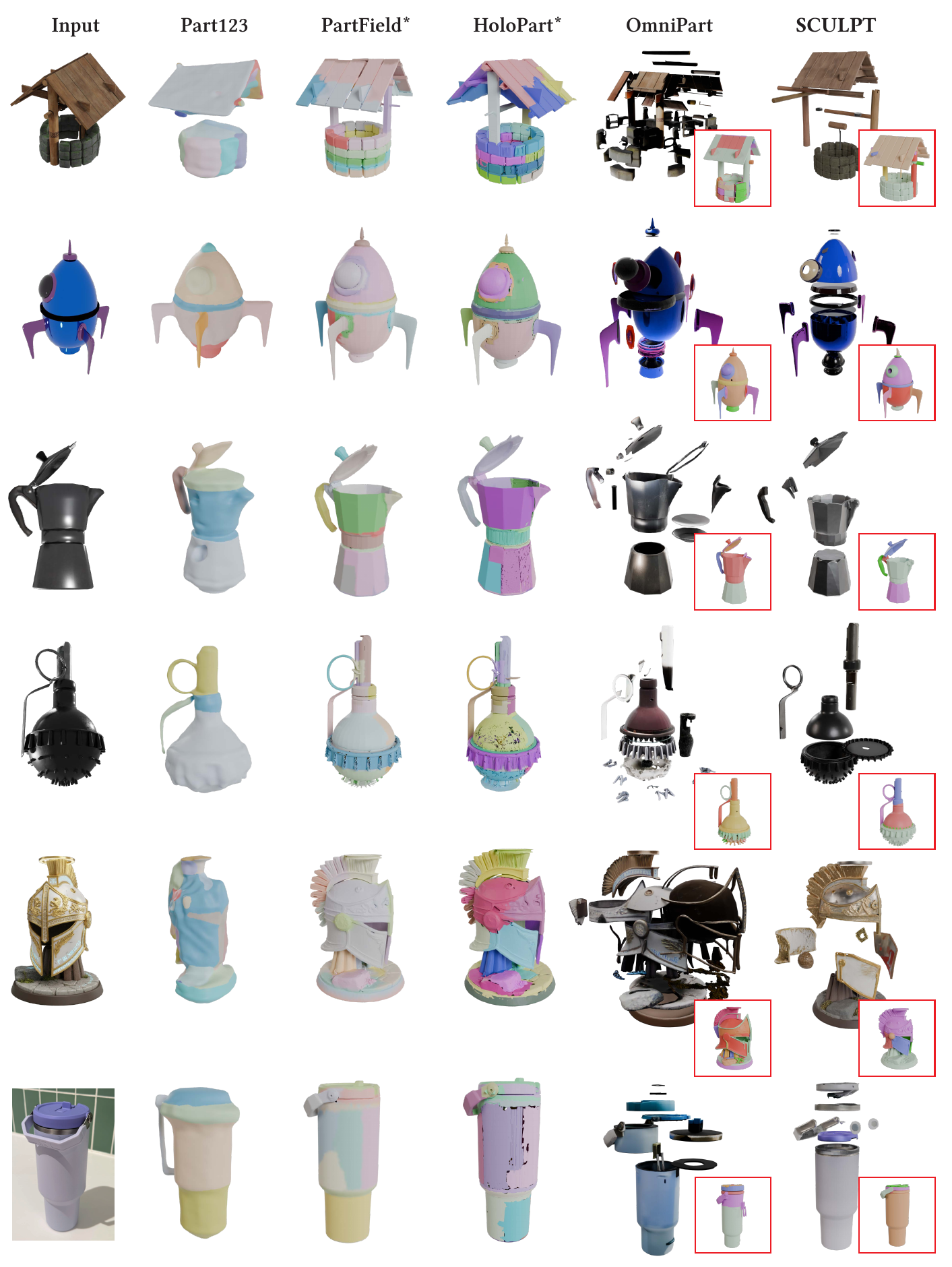}
  \caption{\textbf{Qualitative Comparison.} All methods in each row use the
  same conditioning image. The columns show results from Part123, PartField,
  HoloPart, OmniPart, and SCULPT for four dataset images, one text-to-image
  example, and one real photograph (bottom row). An asterisk marks a baseline rendered with semantic part colors rather
  than generated textures. Red-framed insets show the complete objects obtained
  by reassembling the predicted OmniPart and SCULPT parts in their common object
  frame.}
  \Description{Six qualitative comparisons---four dataset images, one
  text-to-image example, and one real photograph---across Part123,
  PartField, HoloPart, OmniPart, and SCULPT under matched image conditioning.}
  \label{fig:eval_qualitative}
\end{figure}

\paragraph{Evaluation data.}
We evaluate geometry on PartObjaverse, the benchmark released with
SAMPart3D~\citep{sampart3d}. It contains 200 meshes with both instance-level and
semantic part annotations, enabling evaluation of individual parts, semantic
groups, and the reassembled object. The annotations were created independently
of PartVerse-XL, and no labels are transferred between the training and
evaluation sets. We complement this benchmark with the four dataset images,
one text-to-image-generated input, and one real photograph shown in
Fig.~\ref{fig:eval_qualitative} to
examine image-conditioned outputs beyond the benchmark meshes.

\paragraph{Implementation details.}
We initialize the three stage-specific split predictors from the corresponding
public TRELLIS.\,2 checkpoints~\citep{trellis2}; each predictor contains 30
transformer blocks. The stages are trained for 800K optimization steps in total
with AdamW ($\beta_1=0.9$, $\beta_2=0.999$, weight decay $10^{-2}$), a constant
learning rate of $10^{-4}$, and a per-GPU batch size of one on 32 GPUs. We set
the composition-loss weight to $\lambda_{\mathrm{comp}}=0.1$. For the geometry
and material stages, we omit splits whose extracted part contains fewer than 8
or more than 8192 tokens. This removes 1.5\% of splits and 2.3\% of assets from
those two stages; the sparse-structure stage uses the full training set.

\paragraph{Metrics.}
To separate the quality of generated components from the fidelity of their
assembled whole, we evaluate geometry at three levels. At the \emph{part
level}, predicted and reference parts are paired by Hungarian matching and
scored individually. At the \emph{semantic-group level}, each matched
prediction inherits the semantic label of its reference part, and predictions
with the same label are unioned before scoring. At the \emph{object level}, all
predicted parts are assembled and compared with the complete reference mesh.
For every level, we report Chamfer distance (CD) and F1 at thresholds 0.1 and
0.05; lower CD and higher F1 are better, and the 0.05 threshold measures finer
geometric agreement. All meshes are evaluated in the shared $[-1,1]^3$ object
frame.

\paragraph{Compared methods.}
The baselines cover the two main alternatives to subtractive generation. The
first group generates explicit part structure directly and includes
Part123~\citep{part123} and OmniPart~\citep{omnipart}. The second group first
generates a complete object with TRELLIS~\citep{trellis} or
TRELLIS.\,2~\citep{trellis2}, then decomposes it using
PartField~\citep{partfield} or SAM3D~\citep{sam3d}; we additionally evaluate
HoloPart completion~\citep{holopart} where applicable. All methods receive the
same conditioning image when supported. We use official checkpoints and the
inference settings recommended by the respective authors. A dash in
Table~\ref{tab:eval_generation} denotes an unavailable measurement. This
grouping lets us compare SCULPT with both direct part synthesis and post-hoc
decomposition under the same evaluation protocol.

\subsection{Quantitative Comparison}
\label{sec:exp:evaluation}

The central quantitative question is whether a method can improve individual
parts without degrading the object obtained after assembly.
Table~\ref{tab:eval_generation} evaluates both requirements. SCULPT
achieves the lowest CD and the highest F1@.05 at the part, semantic-group, and
object levels, together with the best object-level F1@.1. Its advantage is
therefore not confined to either isolated components or the final union: the
same model remains strong throughout the hierarchy from parts to the complete
object.

\paragraph{Direct part generation.}
OmniPart is the strongest baseline in this group. SCULPT reduces its part-level
CD from 0.0136 to 0.0107 and increases
F1@.05 from 0.7025 to 0.7599. The improvement remains after assembly:
object-level CD decreases from 0.0032 to 0.0020, while F1@.05 increases from
0.8732 to 0.9212. These results show that starting from a complete-object latent
does not merely preserve the initial whole; it also yields more accurate
individual components than direct part generation.

\paragraph{Post-hoc decomposition.}
TRELLIS.\,2+PartField provides the most competitive post-hoc decomposition
baseline and shares the same family of complete-object generator. Relative to
this baseline, SCULPT reduces CD by 7.0\%, 8.5\%, and 4.8\% at the part,
semantic-group, and object levels, respectively, and also improves F1@.05 at
all three levels. TRELLIS.\,2+PartField remains slightly higher on the coarser
part- and semantic-group-level F1@.1 scores (0.8897 vs. 0.8858 and 0.8903 vs.
0.8851). At the stricter 0.05 threshold, however, SCULPT is better at both
levels, consistent with its lower CD and stronger fine-scale geometric
agreement. The object-level gains further show that these more accurate splits
reassemble into a faithful complete shape.

\subsection{Qualitative Comparison}
\label{sec:exp:qualitative}

The benchmark metrics measure geometry, but they do not show whether the
predicted components remain visually meaningful under more varied image
conditions. Fig.~\ref{fig:eval_qualitative} therefore compares the methods on
four dataset images, one text-to-image-generated input, and one real photograph, using the same input within
each row. The figure presents both the separated components and red-framed
reassemblies, making it possible to inspect part structure and complete-object
coherence together.

The examples cover several structures that are difficult to recover as useful
parts. The well combines a broad roof with thin supports and a recessed bucket;
the rocket contains a central body, a nose, and narrow appendages; and the moka
pot requires the lid, handle, body, and base to remain distinct while meeting
at close interfaces. The remaining rows add a ring-shaped handle, a dense
ornament with many small protrusions, and a real tumbler with a lid and handle.
Across these cases, SCULPT separates recognizable components without moving
them out of the coordinate frame of the complete object.

The red-framed insets expose a second requirement that is hidden when parts are
viewed separately: their union should recover the complete asset. SCULPT's
reassemblies retain the silhouette and relative placement of the conditioning
object across both coarse structures and small accessories. This distinguishes
subtractive generation from treating the components as independent objects
that must be normalized or positioned after synthesis. The same property also
explains why the part-level improvements in Table~\ref{tab:eval_generation}
remain visible at the object level rather than being lost during assembly.

Finally, SCULPT decodes geometry and texture for every predicted part, so the
separated outputs retain visual cues such as wood, metal, ceramic, and stylized
color patterns. Asterisks mark baselines rendered with semantic part colors
rather than generated textures. The bottom row shows that the same
part--remainder process can separate the lid, handle, and body of an object
conditioned on a real photograph. Together, the figure complements the
benchmark by showing how geometric fidelity, shared-frame assembly, and
textured part generation appear in the final assets.

\begin{table}[thbp]
  \centering
  \normalsize
  \renewcommand{\arraystretch}{1.08}
  \caption{\textbf{Component ablations on PartObjaverse at the part level.}
  ``Adapt $g,m$'' denotes adaptation of the geometry and material stages,
  $\mathcal{L}_{\mathrm{comp}}$ is composition supervision, and ``Clip'' is
  inference-time clipping to the current remainder. A checkmark enables a
  component and ``--'' disables it. Lower CD and higher F1 are better; bold
  marks the best value.}
  \label{tab:ablations}
  \begin{tabular*}{\linewidth}{@{\extracolsep{\fill}}lcccccc@{}}
    \toprule
    Variant & Adapt $g,m$ & $\mathcal{L}_{\mathrm{comp}}$ & Clip
    & CD$\downarrow$ & F1@.1$\uparrow$ & F1@.05$\uparrow$ \\
    \midrule
    Voxel only & -- & $\checkmark$ & $\checkmark$ & 0.0439 & 0.6801 & 0.5405 \\
    w/o comp. loss & $\checkmark$ & -- & $\checkmark$ & 0.0279 & 0.7379 & 0.5636 \\
    w/o clipping & $\checkmark$ & $\checkmark$ & -- & 0.0260 & 0.7517 & 0.5728 \\
    Full & $\checkmark$ & $\checkmark$ & $\checkmark$
    & \textbf{0.0107} & \textbf{0.8858} & \textbf{0.7599} \\
    \bottomrule
  \end{tabular*}
\end{table}

\subsection{Ablation Studies}
\label{sec:exp:ablation}

Having established the quality of the complete system, we next isolate the
design choices that turn a whole-object generator into a recurrent split
predictor. Table~\ref{tab:ablations} starts from the full model and removes one
capability at a time: adaptation of the downstream geometry and material
stages, composition supervision during training, or support clipping during
inference. All variants are evaluated with the same part-level protocol as the
main comparison.

\paragraph{Adapting the full latent hierarchy.}
The \emph{Voxel only} variant adapts the sparse-structure predictor but leaves
the downstream geometry and material stages unchanged. Its CD increases from
0.0107 to 0.0439, while F1@.1 and F1@.05 fall from 0.8858/0.7599 to
0.6801/0.5405. Sparse support alone is therefore insufficient: the downstream
stages must also adapt to convert the predicted supports into accurate part
geometry.

\paragraph{Constraining the subtractive state.}
The remaining two variants test the training- and inference-time support
constraints. Without composition supervision, CD increases to 0.0279 and
F1@.1/F1@.05 decrease to 0.7379/0.5636, showing that it is important to train
the predicted part and remainder to cover the current object together. Without
support clipping, CD increases to 0.0260 and F1@.1/F1@.05 decrease to
0.7517/0.5728. Allowing a predicted split to extend beyond the current
remainder thus substantially weakens the recurrent decomposition.

Together, these ablations connect the end-to-end gains to the formulation:
full-stage adaptation carries the split through the structured latent
hierarchy, composition supervision trains the two predicted supports to cover
the current object together, and clipping keeps successive predictions within
the current subtractive state.

\section{Conclusion}
\label{sec:conclusion}

We introduced \textbf{SCULPT}, a subtractive formulation of part-aware 3D
generation. Instead of labeling a completed surface or synthesizing a prescribed
collection of separately represented components, SCULPT turns a complete-object
latent into the initial state of a recurrent generative split. Each step jointly
produces one part and the object that remains. This part--remainder transition
provides a fixed prediction signature, an explicit state for later decisions,
and a representation of variable cardinality through the trajectory length.

On PartObjaverse, SCULPT achieves state-of-the-art quantitative performance
among part-structured generation and reconstruction baselines, including the
best CD and F1@.05 at the part, semantic-group, and object levels. Its strong
object-level scores show that fine-grained decomposition does not come at the
expense of the assembled shape. Ablations establish the practical value of
adapting the downstream stages, composition supervision, and inference-time
clipping. Qualitative comparisons further illustrate textured components in a
shared object frame for four dataset images, one text-to-image-generated input,
and one real photograph.

These results identify the explicit remainder state as a useful organizing
principle for part-aware generation: each prediction is conditioned on what
remains to be generated, and the same state provides the termination signal.
Extending this formulation with calibrated stopping, direct granularity control,
and interface-aware objectives is a natural next step.

\clearpage
\bibliographystyle{iclr2027_conference}
\bibliography{references}

\clearpage
\appendix
% \begin{center}
%   {\LARGE\bfseries Appendix}
% \end{center}
\vspace{1em}

\section{Implementation Details}
\label{sec:supp_method}

\subsection{Sparse Packing and Stage Routing}
\label{sec:supp_dft_details}

The implementation preserves the sparse structure, geometry, and material order
of the TRELLIS.\,2 backbone. At the sparse-structure stage, the packed
part--remainder target is represented as a two-channel dense $N^3$ tensor. At
the geometry and material stages, the two outputs generally occupy different
sparse supports. We therefore place their features on the union denoising grid,
retain masks for the part and remainder destinations, and route the jointly
denoised features back to their respective supports. This packing lets the two
outputs share the denoising computation without requiring identical sparse
coordinates.

\subsection{Checkpoint and Control Initialization}

The joint denoising branch is initialized from the corresponding public
TRELLIS.\,2 checkpoint. The residual projections that inject features from the
remainder-control branch are zero-initialized, so the added control path makes
no residual contribution at initialization. The joint and control branches are
then optimized for the part--remainder prediction task.

\subsection{Terminal-State Representation and Pipelining}

For the terminal empty mesh, the sparse-structure target is an all-empty
occupancy grid, while the geometry and material targets are zero-length sparse
tensors. The terminal state therefore uses the same stage-wise predictor
interfaces as nonempty remainders, without a separate terminal output type.
Terminal tuples are included during training so that these representations are
observed by the predictors.

The stage factorization also permits pipelined execution across recurrent
splits. Once step $i$ produces its remainder sparse latent, sparse-structure
prediction for step $i+1$ can begin while step $i$ continues through geometry
and material generation. This schedule changes execution overlap but not the
predicted states or their decoding order.

\section{Additional Details}
\label{app:details}

\subsection{Notation}
\label{app:notation}

Table~\ref{tab:notation} summarizes the main symbols used in SCULPT. We use
subscripts to index subtractive decomposition steps and superscripts
$s\in\{v,g,m\}$ to index the sparse-structure, geometry, and material stages of
the structured latent.

% Keep a notation-only float page aligned with the top of the text block.
\makeatletter
\setlength{\@fptop}{0pt}
\makeatother
\begin{table}[!t]
  \centering
  \caption{\textbf{Notation used in SCULPT.} Hats denote extracted-part
  quantities; for mesh, latent, support, and occupancy pairs, the corresponding
  unhatted quantity denotes the remainder. Initial whole-object states are
  $\mathcal{M}_0$, $\boldsymbol{z}_0$, and $\mathcal{O}_0$.}
  \label{tab:notation}
  \small
  \setlength{\tabcolsep}{4pt}
  \renewcommand{\arraystretch}{1.04}
  \begin{tabular*}{\linewidth}{@{\extracolsep{\fill}}lp{0.72\linewidth}@{}}
    \toprule
    Symbol & Meaning \\
    \midrule
    $\mathcal{I}$ & Conditioning image. \\
    $\mathcal{M}_0$ & Complete normalized training mesh in the shared object frame. \\
    $\hat{\mathcal{M}}_i$ & Mesh of the $i$-th extracted part under the deterministic part order. \\
    $\mathcal{M}_i$ & Remaining mesh after extracting the first $i$ ordered parts; $\mathcal{M}_K=\varnothing$. \\
    $\oplus$ & Mesh-level union in the shared object frame. \\
    \midrule
    $\boldsymbol{z}$ & Generic structured 3D latent. \\
    $\boldsymbol{z}=(\boldsymbol{z}^v,\boldsymbol{z}^g,\boldsymbol{z}^m)$ & Sparse-structure, geometry, and material latent components. \\
    $\boldsymbol{z}_0$ & Initial complete-object latent before any part is extracted. \\
    $\boldsymbol{z}_{i-1}$ & Current remaining-object latent before the $i$-th split. \\
    $\hat{\boldsymbol{z}}_i$ & Extracted part latent predicted at step $i$. \\
    $\boldsymbol{z}_i$ & Updated remaining-object latent after extracting $\hat{\boldsymbol{z}}_i$. \\
    $K$ & Final split index, determined by empty support or the safety cap. \\
    $K_{\max}$ & Maximum number of recurrent splits; $K_{\max}=24$. \\
    $\mathcal{D}$ & TRELLIS.\,2 decoder used to decode each structured latent into a 3D asset. \\
    \midrule
    $P_\theta$ & Joint split predictor, $(\hat{\boldsymbol{z}}_i,\boldsymbol{z}_i)=P_\theta(\boldsymbol{z}_{i-1},\mathcal{I})$. \\
    $P_\theta^s$ & Stage-wise split predictor for $s\in\{v,g,m\}$. \\
    $\boldsymbol{y}_i^s$ & Packed joint target at stage $s$: $\boldsymbol{y}_i^s=(\hat{\boldsymbol{z}}_i^s,\boldsymbol{z}_i^s)$. \\
    $\boldsymbol{y}_i^s(t)$ & Noisy rectified-flow sample, $(1-t)\boldsymbol{y}_i^s+t\boldsymbol{\epsilon}$. \\
    $\boldsymbol{v}_\theta^s$ & Stage-wise velocity predictor for the packed joint target. \\
    \midrule
    $\mathcal{O}$ & Generic hard sparse-structure support map on the O-Voxel grid. \\
    $\mathcal{O}_0$ & Complete-object support associated with $\boldsymbol{z}_0^v$. \\
    $\hat{\mathcal{O}}_i$ & Extracted part support associated with $\hat{\boldsymbol{z}}_i^v$. \\
    $\mathcal{O}_i$ & Remaining support associated with $\boldsymbol{z}_i^v$. \\
    $\cup,\cap,\varnothing$ & Elementwise support union, support intersection, and empty support. \\
    $(\hat{\boldsymbol{p}}_i,\boldsymbol{p}_i)$ & Predicted clean occupancy probabilities for part and remainder. \\
    $\tau_{\mathrm{occ}}$ & Fixed inference-time occupancy threshold. \\
    $N$ & O-Voxel grid resolution, with $\mathcal{O}\in\{0,1\}^{N^3}$. \\
    \midrule
    $\boldsymbol{x}_0$ & Clean target sample used only in the generic flow-matching path. \\
    $\boldsymbol{x}(t)$ & Interpolated flow sample, $(1-t)\boldsymbol{x}_0+t\boldsymbol{\epsilon}$. \\
    $t$ & Flow timestep in $[0,1]$. \\
    $\boldsymbol{\epsilon}$ & Random noise sample for rectified-flow training or sampling. \\
    \midrule
    $\mathcal{L}_{\mathrm{flow}}^s$ & Stage-wise rectified-flow loss on $\boldsymbol{y}_i^s$. \\
    $\mathcal{L}_{\mathrm{comp}}$ & BCE between the previous hard support and the differentiable probabilistic union. \\
    $\lambda_{\mathrm{comp}}$ & Weight for the composition loss. \\
    \bottomrule
  \end{tabular*}
\end{table}

\subsection{Dataset Details}
\label{app:dataset_details}

\paragraph{Evaluation-set exclusion.}
To prevent geometry leakage, we remove any PartVerse-XL training asset whose
SHA-256 identifier matches a PartObjaverse evaluation mesh before filtering and
tuple construction. SAMPart3D and PartVerse-XL are treated as independent
annotation sources: their part vocabularies, boundary placements, and
decomposition granularities may differ, so no PartVerse-XL labels are
transferred to the evaluation benchmark and no SAMPart3D labels are used during
training.

\subsection{Additional Quantitative Results}
\label{app:additional_quantitative}

Table~\ref{tab:category_results} reports the object-class-wise SCULPT results on
PartObjaverse. The class mean is the unweighted average over the eight evaluation
object classes.

\begin{table}[t]
  \centering
  \small
  \caption{\textbf{Semantic-group geometry by object class for SCULPT on
  PartObjaverse.} Lower CD and higher F1 at thresholds 0.1 and 0.05 are better.
  ``Class mean'' is the unweighted mean over the eight listed classes.}
  \label{tab:category_results}
  \setlength{\tabcolsep}{4pt}
  \begin{tabular}{@{}lccc@{}}
    \toprule
    Object class & CD$\downarrow$ & F1-0.1$\uparrow$ & F1-0.05$\uparrow$ \\
    \midrule
    Class mean & 0.0107 & 0.8851 & 0.7614 \\
    \midrule
    Human-Shape & 0.0090 & 0.8979 & 0.7632 \\
    Animals & 0.0116 & 0.8508 & 0.6956 \\
    Daily-Used & 0.0070 & 0.9314 & 0.8399 \\
    Buildings \& Outdoor & 0.0142 & 0.8671 & 0.7411 \\
    Transportations & 0.0091 & 0.8942 & 0.7736 \\
    Plants & 0.0113 & 0.8735 & 0.7360 \\
    Food & 0.0099 & 0.8901 & 0.7977 \\
    Electronics & 0.0136 & 0.8753 & 0.7441 \\
    \bottomrule
  \end{tabular}
\end{table}

\FloatBarrier

\end{document}